\documentclass[letterpaper]{article} 
\usepackage{aaai2026}  
\usepackage{times}  
\usepackage{helvet}  
\usepackage{courier}  
\usepackage[hyphens]{url}  
\usepackage{graphicx} 
\usepackage{natbib}  
\usepackage{caption} 
\usepackage{algorithm}
\usepackage{algorithmic}

\usepackage{newfloat}
\usepackage{listings}
\DeclareCaptionStyle{ruled}{labelfont=normalfont,labelsep=colon,strut=off} 
\floatstyle{ruled}
\newfloat{listing}{tb}{lst}{}
\floatname{listing}{Listing}
\usepackage{multirow}
\usepackage{adjustbox}
\usepackage{booktabs}
\usepackage{enumitem}

\usepackage{amsmath}
\usepackage{amssymb}
\usepackage{subcaption}
\usepackage{xcolor}

\usepackage{changepage}  

\title{\textit{MAMA-Memeia}! Multi-Aspect Multi-Agent Collaboration for Depressive Symptoms Identification in Memes}
\author{
    Siddhant Agarwal\textsuperscript{\rm 1*}, Adya Dhuler\textsuperscript{\rm 2}, Polly Ruhnke\textsuperscript{\rm 1},\\ Melvin Speisman\textsuperscript{\rm 1}, Md Shad Akhtar\textsuperscript{\rm 3*},  Shweta Yadav\textsuperscript{\rm 1*}
}
\affiliations{
    \textsuperscript{\rm 1} University of Illinois at Chicago, USA\\
    \textsuperscript{\rm 2} Creighton University, USA\\
    \textsuperscript{\rm 3} Indraprastha Institute of Information Technology Delhi, India\\
    \textsuperscript{*}Corresponding authors: \{sagarw38, shwetay\}@uic.edu, shad.akhtar@iiitd.ac.in
}

\newcommand{\dataset}{\texttt{RESTOR\textbf{Ex}}}
\newcommand{\llava}{LLaVA 1.5}
\newcommand{\llavanext}{LLaVA-NeXT}
\newcommand{\mcpm}{MiniCPM-V}
\newcommand{\gpt}{GPT-4o}
\newcommand{\model}{\textit{MAMA-Memeia}}

\begin{document}

\maketitle

\begin{abstract}
Over the past years, memes have evolved from being exclusively a medium of humorous exchanges to one that allows users to express a range of emotions freely and easily. With the ever-growing utilization of memes in expressing depressive sentiments, we conduct a study on identifying depressive symptoms exhibited by memes shared by users of online social media platforms. We introduce \dataset\ as a vital resource for detecting depressive symptoms in memes on social media through the Large Language Model (LLM) generated and human-annotated explanations. 
We introduce \model, a collaborative multi-agent multi-aspect discussion framework grounded in the clinical psychology method of Cognitive Analytic Therapy (CAT) Competencies. \model\ improves upon the current state-of-the-art by 7.55\% in macro-F1 and is established as the new benchmark compared to over 30 methods.
\end{abstract}


\section{Introduction}


In the digital age, memes have emerged as a pervasive form of content on online social media platforms, underscoring their escalating significance. In 2020 alone, Instagram, a popular social media app, reported at least one million posts shared every day mentioning the word ``meme" \citep{bbcSurprisingPower}. Beyond entertainment, memes have grown to become an important outlet for expressive posting as well \citep{10.1145/3359170}, as people have increasingly begun to use memes to communicate their emotional struggles, most particularly depression, where a simple Google search for ``depression memes" shows more than $84,700,000$ results. 
\citet{Akram2020} observe that sharing humor-intended depressive memes can be beneficial for some individuals, as presenting a negative experience humorously may create a sense of peer support among viewers who have gone through similar situations. 
The importance of the visual modality is further cemented considering that the majority of social media posts today contain image or video content which increases engagement by around 100\% \cite{london_research_nodate}.

Recognizing the importance of memes in online social media today, meme analysis research has gained traction, starting with the Facebook Hateful Memes Challenge \citep{kiela_hateful_2020}. However, research in the area has been mostly focused on aspects such as harmfulness \citep{pramanick-etal-2021-detecting} and cyber-bullying \citep{ jha-etal-2024-meme}.
In this work, we work on a relatively understudied area in meme analysis and apply it to the mental health domain. We work to identify seven fine-grained depressive symptoms 
based on the clinically established 9-scale Patient Health Questionnaire (PHQ-9) \citep{kroenke_phq-9_2001}.

\begin{figure}[t!]
    \centering
    \includegraphics[width=\columnwidth]{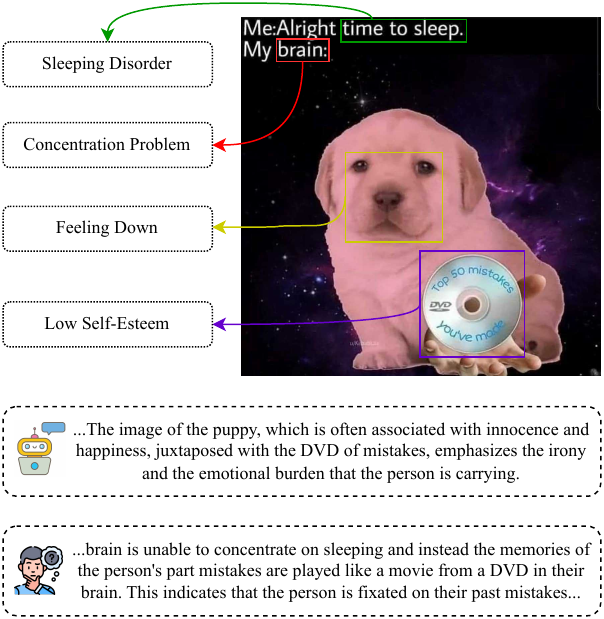}
    \caption{An example of a depressive meme from \dataset\ with labeled fine-grained depressive symptoms and \textit{explanations} from the  LLM agent and Human}
    \label{fig:intro-example}
\end{figure}

The task involves multi-label classification for the seven identified symptoms in the multimodal memes as proposed by \citet{Yadav-etal-2023-towards}. We derive a new dataset, \dataset, consisting of \textit{explanations} generated by state-of-the-art Multimodal Large Language Models and supplemented with human-annotated explanations for the memes. 
For this task, we examine the application of Multimodal Large Language Models in understanding complex multimodal data of memes in the context of mental health. This is a complex task for language models given the mix of sarcasm, irony and other types of figurative speech involved in memes.
For example, in Figure \ref{fig:intro-example}, the meme has the caption \textit{``Me: Alright time to sleep."}, \textit{``My brain:"}, with a picture of a puppy holding a DVD that says, \textit{``Top 50 mistakes you've made"}. 
The human annotator emphasizes on the meme author's fixation on their past mistakes, correctly identifying the \textit{`Low Self-Esteem'} symptom. Further, the 
LLM-generated explanation, recognizing the image of the sad expression of the puppy (often used in a positive context) juxtaposed with negative emotions relating to the DVD, 
understands the irony of the situation to correctly assign the \textit{`Feeling Down'} label to this meme. 
Overall, through a multi-dimensional understanding of the various figurative speeches in the meme, the method should capture all four depressive symptoms expressed by the meme author -- \textit{Low Self-Esteem}, \textit{Concentration Problem}, \textit{Sleeping Disorder}, and \textit{Feeling Down}.

In this work, we introduce \model, a Multi-Aspect Multi-Agent collaborative framework for identifying depressive symptoms from memes. We ground this proposed methodology in clinical psychology by adapting the Cognitive Analytic Therapy (CAT) Competencies \cite{parry_developing_2021} into our Multi-Aspect prompting setup which is used by medical professionals for analyzing thinking patterns. We supplement our analysis with extensive comparative results establishing the efficacy of \model\ as the preeminent method for the task of depression symptom analysis.

Our contributions are summarized as follows- 
\begin{itemize}
    \item \textbf{Novel Dataset:} Introducing the \dataset\ dataset, with LLM-generated explanations and human annotated gold label explanations.
    \item \textbf{Novel Methodology:} Introducing the \model\ framework, a novel state-of-the-art multi-agent discussion framework that builds upon the foundations of clinical psychology for multi-aspect prompting.
    \item \textbf{In-depth Human Evaluation:} Conducted a human evaluation with domain experts to address key research questions regarding LLM-generated content across five aspects. 
\end{itemize}

\section{Related Work}

\begin{figure*}[t!]
    \centering
    \begin{subfigure}[t]{0.3\textwidth}
        \centering
        \includegraphics[height=0.8in]{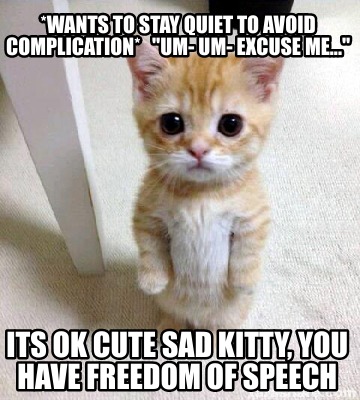}
        \caption{Misclassifications in Test Samples: Original- [FD, LOI], Updated- [FD, LSE] }
        \label{fig:misclass}
    \end{subfigure}%
    \hfill
    \begin{subfigure}[t]{0.3\textwidth}
        \centering
        \includegraphics[height=0.8in]{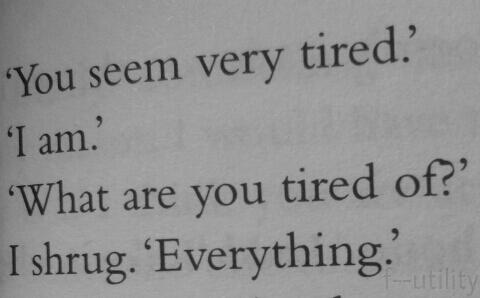}
        \caption{Non-meme Images: image that corresponds to a quote }
        \label{fig:non-meme}
    \end{subfigure}
    \hfill
    \begin{subfigure}[t]{0.27\textwidth}
        \centering
        \includegraphics[height=0.8in]{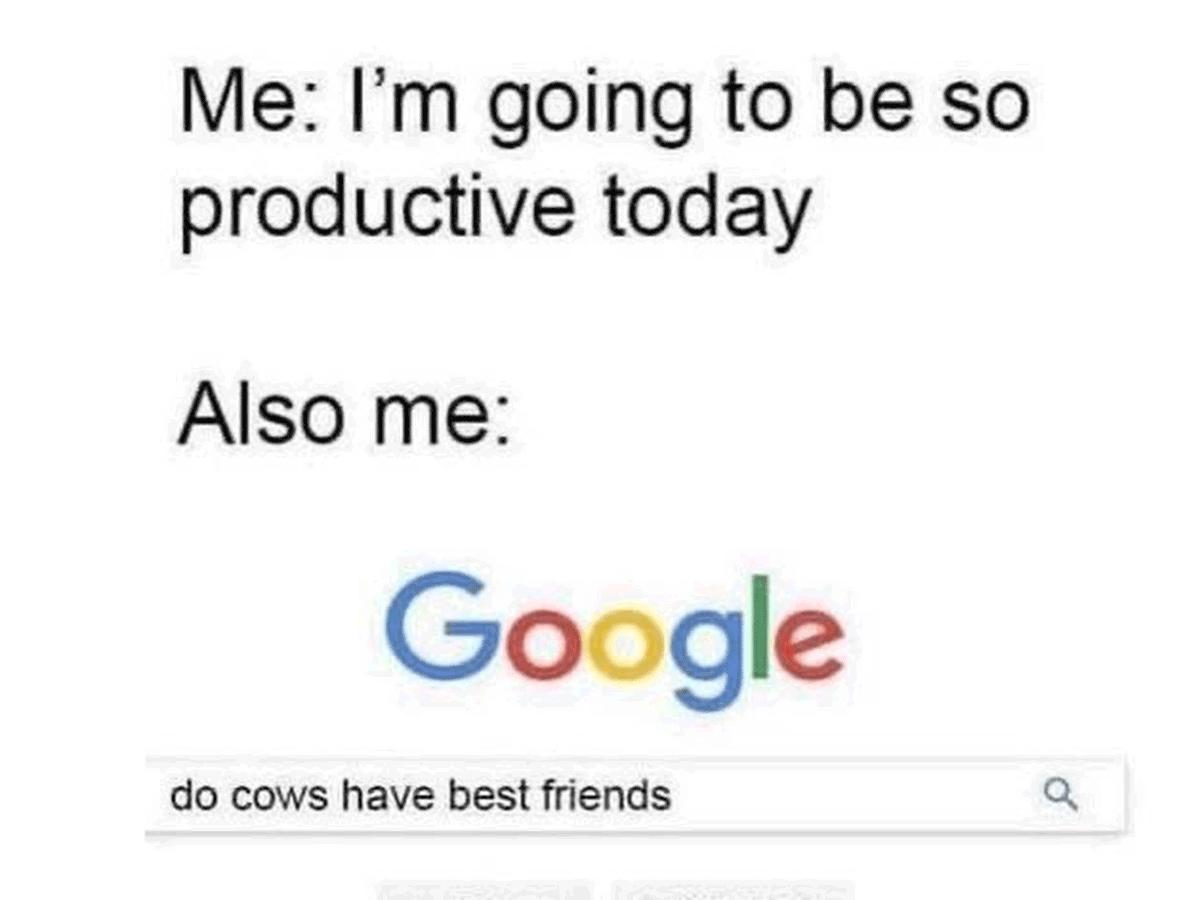}
        \caption{Inaccurate `Lack of Energy' samples}
        \label{fig:lackofenergy}
    \end{subfigure}

    \caption{Examples of meme samples from \texttt{RESTORE} that were re-annotated for the curation of \dataset}
\end{figure*}

\label{sec:related_works}
In recent years, efforts towards the analysis of memes and subsequent datasets in the area have been focussed mainly on aspects such as hatefulness \citep{kiela_hateful_2020}, harmfulness \citep{Sharma_Agarwal_Suresh_Nakov_Akhtar_Chakraborty_2023}, 
and emotions \citep{sharma-etal-2020-semeval, DBLP:conf/defactify/MishraSPCRRCDSC23}. Recently, \citet{JoshiSaurav2024CIMA} attempted to understand the propagation of meme content across social media platforms.

While most work on modeling for meme analysis has focused on contrastive learning setups \citep{mei-etal-2024-improving} and knowledge fusion techniques \citep{sharma-etal-2023-memex}, the recent direction of the area has been towards the use of LLMs \citep{jha-etal-2024-memeguard}. Works such as those by \citet{agarwal-etal-2024-mememqa} and \citet{zhong_multimodal_2024} have explored the application of explanations generated by LLMs for explaining memes in various contexts. However, rigorous analysis of the generated content remains a challenge that we attempt to resolve with this work.

While research on memes expanded, their application in the mental health domain remained limited. \citet{Yadav-etal-2023-towards} extended meme analysis research towards the mental health domain with the release of the \texttt{RESTORE} dataset for detecting fine-grain depressive symptoms in memes. \citet{mazhar2025figurativecumcommonsenseknowledgeinfusionmultimodal} expand this research by utilizing a knowledge-fusion and retrieval framework for depression symptom identification as well as introducing a dataset for detecting anxiety symptoms. We aim to expand on these works with a more elaborate method by utilizing LLM agents.

Previous work on depressive symptom analysis has focused mostly on textual data such as Tweets and Reddit posts, with limited work on multimodal content such as memes \citep{Yadav-etal-2023-towards}. Works such as \citet{yadav-etal-2021-when, yadav-etal-2018-multi, benton-etal-2017-multitask} have focused on analyzing depressive symptoms by collecting data from online social media platforms, mostly on tweets of users for detection or tracking of depression symptoms in users. Works such as \citet{yates-etal-2017-depression, yadav2020identifying} have served as important benchmarks for analyzing textual depressive data. Typically text-based depression symptom analysis works utilize simpler BERT-based methods with specialized modules \cite{ZHANG2023103417}.  Recently, \citet{wang-etal-2024-explainable} analyzed the utility of LLMs for estimation of the levels of depression by prompting LLMs with a questionnaire. \citet{chen-etal-2024-depression} have utilized LLMs for synthetic data creation in the form of interviews of depressed users highlighting the capabilities of LLMs to work in the mental health context. This work aims to extend depression analysis to the multimodal setting and applying advanced LLM-based methods, including multi-agent methods which have not yet been effectively utilized in depression related tasks.

\begin{table}
    \caption{Class-wise distribution of fine-grained depression symptom labels in \dataset.}
    \label{tab:dataset}
\begin{adjustbox}{width=\columnwidth}
    \begin{tabular}{c | c c c c c c c | c}
    \toprule
      & LOI & FD & ED & SD & LSE & CP & SH &  Total\\
    \midrule
    \textit{Train} & 441 &  1781& 1714&997&703&348 &1357& 7096\\
    \textit{Test}& 66 &  219& 85&78&114&73 &82&520\\
    \textit{Validation} & 37 &130&54 &36& 79& 28&54& 310\\

    \bottomrule
    \end{tabular}
\end{adjustbox}
    
\end{table}

\section{The \dataset\ Dataset}
We introduce \dataset
, a dataset for the multi-label classification of seven fine-grained PHQ-9 depressive symptom categories -- \textit{Feeling Down (FD), Lack of Interest (LOI), Eating Disorder (ED), Sleeping Disorder (SD), Concentration Problem (CP), Low Self Esteem (LSE), and Self Harm (SH)}. We provide expert annotated human explanations and conduct extensive expert analysis of all generated content. 
Table \ref{tab:dataset} provides label-wise distribution of \dataset.

\subsection{Dataset Curation and Filtering}
The \dataset\ dataset is derived from the \texttt{RESTORE} dataset, which collects meme images from the social media platforms, Twitter and Reddit. The dataset consists of human annotations for eight fine-grained depression symptom categories in the test and validation subsets of the dataset. The training subset is supplemented with automatically curated samples using keyword-based search (such as `eating disorder memes', `feeling down memes', etc.).

In the curation of the \dataset\ dataset,
we identify and correct the following inconsistencies in the original \texttt{RESTORE} dataset:
\textbf{(i)} Re-annotated the test and validation sets of the dataset with additional annotators to mitigate issues with misclassified labels as shown in Figure \ref{fig:misclass}.
\textbf{(ii)} Removed non-meme images such as quotes from the 
 dataset as shown in Figure \ref{fig:non-meme}. leading to a reduction of about 20\% in the dataset size.
 \textbf{(iii)} Removed the \textit{`Lack of Energy'} label due to the lack of accurate training labels as shown in Figure \ref{fig:lackofenergy}. This was done as all of the samples for this class in the training set (471 samples) were automatically curated with significant issues.

\subsection{Explanations in \dataset}

To enhance the utility of \dataset\ as a resource, we introduce ground-truth human-written explanations for the test subset (520 samples). This ground-truth explanation serves as a resource for future research to compare model-generated explanations with a human-annotated dataset. 
An \textit{explanation} by the human-annotators, through their nuanced thought process, encapsulates important aspects such as figurative language (for example, sarcasm and metaphors), commonsense reasoning, and is grounded in their cultural awareness which allows explanations to be aware of the cultural phenomena of the time. These detail-rich explanations serve as a textual description of the cross-modal information in a meme image.

\subsection{Annotation Guidelines}
The curation of the \dataset\ dataset consists of two annotation tasks for dataset re-annotation and filtering, and for curation of human-written gold label explanations.
\newline
\textbf{Task 1}: For a given meme, the annotators are required to determine one or more of the 
seven depressive symptoms conveyed in the meme 
    according to PHQ-9 definitions \citep{kroenke_phq-9_2001}. Any non-meme image is filtered in this process. This annotation is performed by two annotators trained on the task of fine-grained depression symptom identification as per the guidelines of the original dataset curation \cite{Yadav-etal-2023-towards}. The inter-annotator agreement was captured as the Krippendorff's Alpha coefficient \cite{krippendorff_measuring_2004}, a metric widely used for measuring reliability in annotations for multi-label tasks. The coefficient is obtained as 0.833 using MASI distance \cite{passonneau-2006-measuring} as the distance function, representing strong-agreement between the annotators. 
\newline
\textbf{Task 2}: For each meme, the annotators are required to provide the human-annotated ground-truth explanations in the provided text field. The annotator is instructed to capture their thought process and incorporate details such as the use of figurative language, commonsense knowledge, cultural references, and visual cues in their provided explanations. This annotation process is performed by two domain-expert annotators.

\subsection{LLM Generated Explanations}
We propose to extend the dataset further with Multimodal LLM generated explanations as a proxy for human-written annotations for large datasets where human annotation is not feasible such as for the over 7000 samples in the train dataset. We generate explanations using three open-source and three closed-source LLMs 
and conduct a human analysis on these explanations to evaluate their utility as a proxy for human-written explanations.

\paragraph{Analysis on LLM Generated Explanations}
\label{sec:human_analysis}

We focus on five research questions in order to comprehensively judge the quality of LLM generated explanations: \textbf{RQ1:} \textit{Are LLMs fluent?}, \textbf{RQ2:} \textit{Are generated explanations relevant?}, \textbf{RQ3:} \textit{Do LLMs capture figurative meaning?}, \textbf{RQ4:} \textit{Are LLMs persuasive?} and \textbf{RQ5}: \textit{Understanding the appeal of some explanations}.
We conduct this analysis on six selected model explanations and human-annotated explanations based on a detailed questionnaire on a subset of test samples. We choose three open source models -- \llava, \llavanext, \mcpm and three closed source models -- \gpt, Claude 3.5 Sonnet and Gemini-2.0-flash. The results of this human analysis clearly establish the superiority of closed-source LLMs such as Gemini \cite{geminiteam2024geminifamilyhighlycapable} over open-source models for the task of \dataset\ motivating our decision of utilizing such closed-source models as the backbone of our methodology.

\section{Methodology}
\label{sec:methodology}

\begin{figure*}
    \centering
    \includegraphics[width=0.8\textwidth]{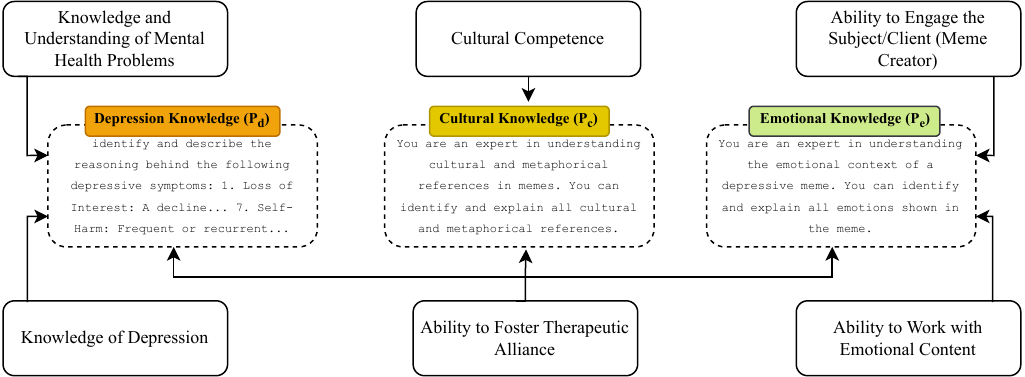}
    \caption{Cognitive Analytic Therapy (CAT) Competencies \citep{parry_developing_2021} adapted as guidelines form the basis for Multi-Aspect Prompting. Multiple criteria combine to form each Knowledge Aspect. We design three Aspect-specific prompts -- $P_d$ (Depression Knowledge), $P_e$ (Emotional Knowledge), and $P_c$ (Cultural Knowledge) -- to generate the corresponding explanations -- $E_d$, $E_e$, and $E_c$.  We use these explanations individually and concatenated as $<E_d , E_e , E_c>$ for our experiments.}
    \label{fig:cat-eval}
\end{figure*}

In this section we propose the \model\ framework, a collaborative multi-agent discussion framework that utilizes clinical psychology principles to detect depressive symptoms. We first describe the domain expert's perspective on the task of depressive symptom classification and then present our approach to integrating this perspective in a multi-agent setup through multi-aspect prompting.

\subsection{Domain Expert Perspective}
\label{sec:domain}

The psychology underlying meme with mental health is centered around emotional expressions and coping mechanisms (particularly, in the form of humor). Specifically, meme allows individuals to express complex emotions surrounding their mental health in a relatable manner through their hypernarrativity \citep{wagener2021postdigital}. They provide a platform for communal discourse and validation for individuals in similar situations by supporting a larger cultural framework to help process individual experiences \citep{wagener2021postdigital}.

Humor can be used to create emotional distance and to help people process feelings. This enables them to cope with stress, depression, and anxiety, especially during important developmental transitions \citep{sarink2023humor,erickson2006adolescent}. While positive humor (affiliative, self-enhancing) is associated with stable psychological adjustment, negative humor (especially self-defeating humor seen in memes) is associated with higher depressive symptoms \citep{erickson2006adolescent}. Therefore, although it helps tops to build social relationships and navigate depressive symptoms, it harms self-esteem in the process \citep{erickson2006adolescent}. Specific to memes, humor provides an avenue for emotional catharsis by channeling dark or uncomfortable topics toward humor \citep{sarink2023humor}.


 \paragraph{Cognitive Analytic Therapy for Understanding Depressive Memes}
Cognitive Analytic Therapy (CAT) is a form of talking therapy that focuses primarily on identifying patterns of thinking, feeling, and behavior \cite{nhs_cognitive_nodate}. It is widely used in the clinical therapy setting for people living with depression, anxiety, or eating problems, who self-harm or have personal or relationship problems.
In this study, we adapted CAT - General Therapeutic Competencies \citep{parry_developing_2021} to formulate evaluative guidelines for interpreting memes on mental health themes. It is highly applicable to the task due to its structured approach to categorizing and assessing both larger psychological patterns and more subtle emotional nuances and risk assessments. We leverage 
eight CAT criteria focusing on meme understanding -- (i) \textit{Knowledge and Understanding of Mental Health Problems}, (ii)\textit{ Knowledge of Depression}, (iii) \textit{Cultural Competence}, (iv) \textit{Ability to Engage the Subject/Client (Meme Creator)}, (v) \textit{Ability to Foster Therapeutic Alliance}, (vi) \textit{Ability to Work with Emotional Content}, (vii) \textit{Ability to Assess and Manage Risk of Self-Harm}, (viii) \textit{Knowledge of Ethical Guidelines}.

\subsection{Multi-Aspect Prompting}
\label{sec:multi-aspect}

\begin{figure*}
    \centering
    \includegraphics[width=0.9\textwidth]{model_diag.pdf}
    \caption{Overview of the \model\ framework with Gemini-2.0-flash\;\includegraphics[height=8pt]{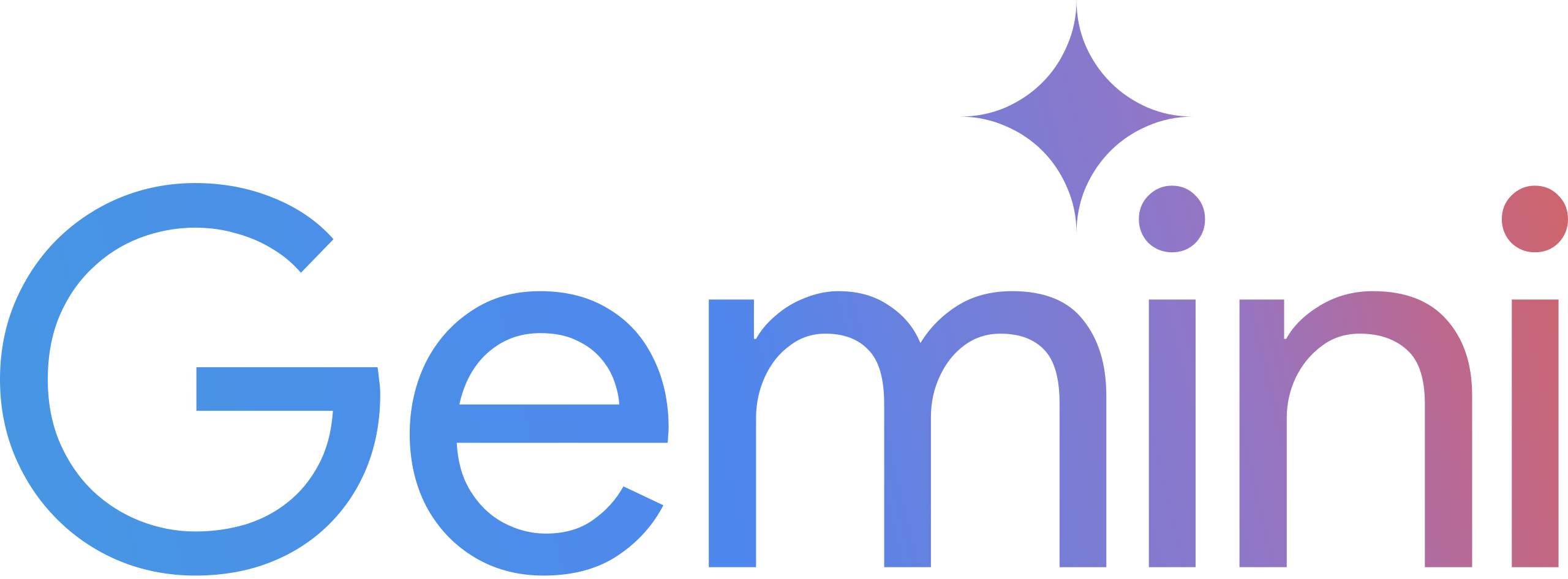}, Claude 3.5 Sonnet\;\includegraphics[width=8pt]{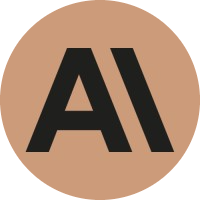}, GPT-4o\;\includegraphics[width=8pt]{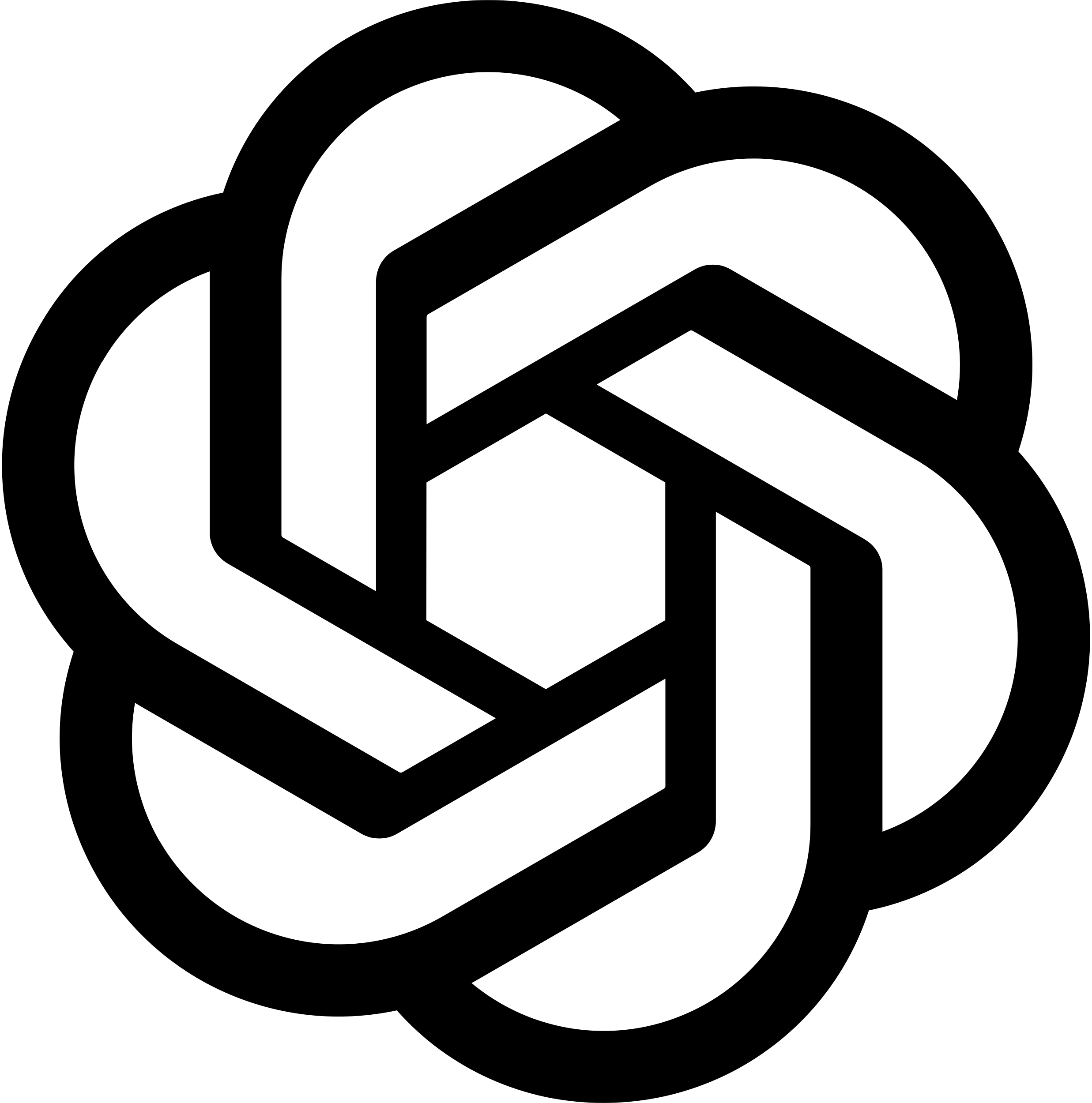}, consisting of three phases- (1) \textit{Independent Ideation}: Agents generate their initial predictions and reasoning, (2) \textit{Collaborative Discussion}: Agents collaborate to discuss and reconsider their predictions, (3) \textit{Consensus Resolution}: Weighted vote on final confidence estimates to determine predicted labels.}

    \label{fig:model}
\end{figure*}

We ground our approach in the CAT criteria described in section \ref{sec:domain}. These guidelines highlight three major aspects of knowledge required for understanding depressive memes: 
\newline\textbf{(i) Depression Knowledge} -  This aspect deals with providing knowledge about the specific depressive symptoms to the model. We provide this knowledge with the use of definitions for the seven symptoms as laid out by \citep{Yadav-etal-2023-towards}. These definitions ensure that the model avoids misinterpretation of the depressive symptoms. 
\newline\textbf{(ii) Emotional Knowledge} - A key consideration while determining depressive symptoms according to the CAT guidelines is the consideration of the emotional state of the user. To this effect, we include this knowledge aspect to ensure that the model considers the emotional states that may be associated with depressive themes in the meme.
\newline\textbf{(iii) Cultural Knowledge} - This knowledge aspect aims to provide information about important cultural phenomena such as pop culture references or metaphorical interpretations that form a key part of meme understanding. With the inclusion of such knowledge it is expected that the model would be able to recognize the figurative language of the meme across the textual and visual modalities.

Based on these, we design three Aspect-specific prompts $P_d$ (Depression Knowledge), $P_e$ (Emotional Knowledge), and $P_c$ (Cultural Knowledge) as described in Figure \ref{fig:cat-eval}.

\subsection{\model: A Multi-Agent Multi-Aspect Inference Framework}

We introduce the \model\ framework, which builds upon the Multi-Aspect Prompting setup by utilizing multiple agents that capture specific knowledge of depressive memes in the prediction of the final labels. Our approach is inspired from multi-agent works \citep{wu2024autogen, 10.5555/3692070.3692537, chen-etal-2024-reconcile}
which highlights the efficacy of multi-agent setups. 

We utilize this knowledge in combination with our study of Multi-Aspect Prompting by deploying three recent state-of-the-art LLMs - GPT-4o \cite{openai2024gpt4ocard}, Claude-3.5 Sonnet \cite{claude} and Gemini-2.0-flash \cite{geminiteam2024geminifamilyhighlycapable} - in a multi-agent setup. Our choice of these models is based on the human analysis provided in subsection \ref{sec:human_analysis}. The \model\ framework consists of three phases: \textit{Independent Ideation, Collaborative Discussion and Consensus Resolution}. It is demonstrated in Figure \ref{fig:model} and described in detail below for $n$ agents.

\paragraph{\textbf{Independent Ideation}} 
The first step in the \model\ framework is the generation of initial symptom predictions and reasoning for the given meme from each agent.
Formally, given a list of agents $A = \{A_1, A_2, ...,A_n\}$ of size $n$,
each agent $A_i \in A$ generates an initial prediction, $p_{i}^{(0)}$, which is a list of $s_i$ symptoms predicted by it. 
The study \cite{xiong2024can} has shown the effectiveness of LLM-derived confidence scores, therefore, we also generate confidence estimates, $c_{i,j}^{(0)} \in [0, 1] \;\forall\; j \in s_i$.
Each agent, $A_i$, also generates an initial explanation, $e_i^{(0)}$ that explains the thought process behind it's predictions. To achieve this, $A_i$ is designated a Aspect-Specific System Prompt, $P_{a,i} \in \{P_d, P_e, P_c\}$, for each agent to focus on one knowledge aspect.  
Each agent $A_i$ is then prompted with the Aspect-specific Prompt, $P_{a,i}$, the meme image, $M_{img}$, and the user prompt, $P_{usr}$ (detailed prompts in suppl.). For each agent, $A_i$, this can be formulated as:
\begin{equation}
    \begin{aligned}
    p_{i}^{(0)}, c_i^{(0)}, e_i^{(0)} = A_i({P}_{a,i}, M_{img}, P_{usr}) \\
    \end{aligned}
\end{equation}

\paragraph{\textbf{Collaborative Discussion}}
In the next phase of our framework, all agents undergo $R$ rounds of collaborative discussion. This is done to ensure each agent is fairly exposed to the thought process of other agents before coming up with their final predictions. For each discussion round, $r$, every agent is provided a Discussion prompt, $P_{dsc}$, which has multiple characteristics. The discussion prompt first introduces the agents to the idea that they are in a collaborative environment by asking them to review the responses of other agents and then reconsider their response. We call this the initiator prompt, $P_{ini}$. Secondly, the discussion prompt for discussion round, $r$, consists of the responses of the other agents for round, $r-1$ (for $r = 1$ this refers to the Independent Ideation phase). Each agent is provided with both explanation, $e_j^{(r-1)}$, as well as predictions, $p_j^{(r-1)}$, generated by the other agents, $A \setminus A_i$. This is supplemented with the confidence estimates, $c_j^{(r-1)}$, for each prediction in $p_j^{(r-1)}$. The availability of all three components, explanation, predictions, and confidence estimates, allows the agent to condition on the reasoning and confidence associated with the predictions before it reconsiders its initial judgement. The round $r$ discussion prompt, for agent $A_i$, is:
\begin{center}
     $P_{dsc, i}^{(r)} = P_{ini} \;+ \big\|_{j=1, j \neq i}^{n} \{ e_j^{(r-1)}, p_j^{(r-1)}, c_j^{(r-1)} \}$
\end{center}
This discussion prompt is then passed along with the entire conversation history of the agent, $h_i^{(r)}$, to obtain the predictions, confidence estimates and explanation of the agent for round $r$ as:
\begin{equation}
    p_{i}^{(r)}, c_i^{(r)}, e_i^{(r)} = A_i(P_{dsc, i}^{(r)}, h_i^{(r)})
\end{equation}

\paragraph{\textbf{Consensus Resolution}} At the end of $N = R + 1$ rounds, we determine the final predictions through our consensus resolution algorithm which performs a weighted vote between the agents based on the confidence estimates. For a given threshold, $t$, the final labels, $L$, are determined as follows:
\begin{equation}
L= \left[\,  j \;\middle|\; (\frac{1}{n}\sum_{i=1}^{n} c_{i,j}^{(N)}\,\mathbb{I}\{j\in p_i^{(N)}\} ) > t \,]
\right]
\end{equation}


The consensus resolution algorithm is used in order to eliminate labels for which the agents have low confidence. It also ensures that there is a balance between the agents and extremely high confidence of one agent does not bias \model.

\begin{table}
    \centering
    \caption{Comparison of \model\ with unimodal and multimodal baselines along with previous state-of-the-art. 
    }
    \label{tab:benchmarking}
    \resizebox{0.98\columnwidth}{!}{
    \begin{tabular}{lcc}
    \toprule
         Model & Macro-F1 & Weighted-F1 \\
         \midrule
         \textbf{Unimodal Text} & & \\
         \quad \textit{OCR:} & & \\
         \qquad BERT \citep{devlin-etal-2019-bert}& 62.02 & 62.63 \\
         \qquad MentalBERT \cite{ji-etal-2022-mentalbert}& 63.77 & 64.47 \\
         \qquad BART \cite{lewis-etal-2020-bart}& 44.71 & 49.46 \\
         \qquad MentalBART \cite{yang2023mentalllama}& 61.76 & 62.43 \\

         \quad \textit{Explanation:} & & \\
         \qquad BERT& 63.39 & 63.96 \\
         \qquad MentalBERT & 64.62 & 65.27\\
         \qquad BART & 55.81 & 57.11\\
         \qquad MentalBART & 64.96 & 64.30\\
        \midrule
        \textbf{Unimodal Image} & & \\
        \qquad ViT \cite{dosovitskiy2021an}& 34.96 & 39.04 \\
        \qquad ResNet \cite{He2015DeepRL}& 27.14 & 33.46 \\
        \qquad EfficientNet \cite{tan_efficientnet_2019}& 25.18 & 31.61 \\
        \midrule
        \textbf{Multimodal} & & \\
        \quad \textit{Image + OCR} \\
         \qquad CLIP \cite{radford2021learningtransferablevisualmodels}& 45.83 & 48.09 \\
         \qquad VisualBERT \cite{li2019visualbertsimpleperformantbaseline}& 62.70 & 63.67 \\
         \qquad ViT + BERT& 38.57 & 42.24 \\
        \quad \textit{Image + Explanation} \\
        \qquad CLIP & 39.23 & 42.56\\
        \qquad VisualBERT & 62.37 & 62.25 \\
        \qquad ViT + BERT & 37.78 & 40.58\\
         \quad Previous SOTA\\ 
         \qquad\citet{Yadav-etal-2023-towards} ${\dagger}\;$ & 65.18 & 64.67 \\
         \midrule
         \model\;\includegraphics[height=8pt]{Google_Gemini_logo.svg.png} \;\includegraphics[width=8pt]{Anthropic-AI.png}\;\includegraphics[width=8pt]{openai.png} 
         & \textbf{72.73} & \textbf{72.45}\\
         \midrule
         $\Delta_{\model \;\textbf{-}\; \dagger}$ & {\color{blue} $\uparrow 7.55\%$}  & {\color{blue} $\uparrow 7.78\%$}\\
         \bottomrule
    \end{tabular}
     }
\end{table}




\section{Experiments: Benchmarking \dataset}
\label{sec:results}




\paragraph{Baselines} To establish the effectiveness of the \model\ framework, we first compare its performance to several relevant unimodal and multimodal baseline architectures as described in Table \ref{tab:benchmarking}. These baselines use the following models:
\textbf{(i)} \textbf{Unimodal Text}: BERT, MentalBERT, BART, MentalBART
\textbf{(ii)} \textbf{Unimodal Image}: ViT, ResNet, EfficientNet
\textbf{(iii)} \textbf{Multimodal}: CLIP, VisualBERT, BERT + ViT, SOTA \citep{Yadav-etal-2023-towards}

\begin{table*}[h!]
\centering
\caption{Comparison of \model\ with Mulitmodal LLM based single-agent setups and Human-annotated explanations. \model\ is the best performing method in macro, micro and weighted F1 scores. Vanilla LLM Explanations refer to the prompting setup without the inclusion of CAT knowledge based prompts. FD: Feeling Down, LOI: Lack of Interest, SH: Self-Harm, ED: Eating Disorder, LSE: Low Self-Esteem, CP: Concentration Problem, SD: Sleeping Disorder. }
\label{tab:results}

\resizebox{0.9\textwidth}{!}
{%
\begin{tabular}{lcccccccccc}
\toprule
{Method} & FD & LOI & SH & ED & LSE & CP & SD & Macro-F1 & Micro-F1 & Weighted-F1\\ 
\midrule
\textbf{Human Explanations} & 63.00	&47.00	&77.00	&72.00	&54.00	&57.00	&81.00 &65.00	&62.00	&64.00\\
\midrule
\textbf{Vanila LLM Explanations} \\
\quad\textit{Open-Source LLMs} \\
\qquad \llava\  & 55.00	&37.00	&34.00	&48.00	&38.00	&47.00	&67.00	&47.00	&48.00	&48.00\\
\qquad \llavanext\ & 62.00	&43.00	&60.00	&71.00	&46.00	&53.00	&79.00	&59.00	&57.00	&59.00\\
\qquad \mcpm\  & 62.00	&40.00	&53.00	&73.00	&51.00	&53.00	&78.00	&59.00	&58.00	&59.00\\
\quad\textit{Closed-Source LLMs} \\
\qquad GPT-4o \;\includegraphics[width=8pt]{openai.png}&
63.00	&39.00	&58.00	&73.00	&46.00	&54.00	&86.00	&60.00	&57.00	&60.00\\
\qquad Cluade 3.5 Sonnet \;\includegraphics[width=8pt]{Anthropic-AI.png}&
70.78	&48.25	&77.71	&81.01	&54.35	&57.92 &88.76	&68.39	&66.19 &68.74\\
\qquad Gemini-2.0-flash \;\includegraphics[height=8pt]{Google_Gemini_logo.svg.png}&
63.09&32.02	&75.86	&75.47	&49.22	&60.47	&80.22 &62.34	&58.19	&62.55\\
\midrule
\textbf{Multi-Aspect Prompting} \\
\quad \textit{Depression Knowledge ($P_d$)} \\
\qquad GPT-4o \;\includegraphics[width=8pt]{openai.png}&
65.13&\textbf{53.33}	&79.52	&78.21	&45.02	&58.46	&86.22 &66.56	&62.67	&65.66\\
\qquad Cluade 3.5 Sonnet \;\includegraphics[width=8pt]{Anthropic-AI.png}&
68.27&45.45	&77.30	&82.93	&48.46	&60.09	&84.71 &66.74	&63.60	&66.75\\
\qquad Gemini-2.0-flash \;\includegraphics[height=8pt]{Google_Gemini_logo.svg.png}&
66.40&46.60	&73.85	&74.21	&55.86	&65.92	&78.95 &65.97	&64.93	&65.99\\
\quad \textit{Cultural Knowledge ($P_c$)} \\
\qquad GPT-4o \;\includegraphics[width=8pt]{openai.png}&
69.55&49.43	&69.50	&79.22	&49.87	&60.60	&86.06 &66.31	&64.67	&66.60\\
\qquad Cluade 3.5 Sonnet \;\includegraphics[width=8pt]{Anthropic-AI.png}&
71.09&45.57	&72.99	&83.23	&44.95	&56.67	&85.21 &65.67	&62.61	&66.31\\
\qquad Gemini-2.0-flash \;\includegraphics[height=8pt]{Google_Gemini_logo.svg.png}&
66.77&38.96	&61.88	&75.74	&45.55	&53.46	&76.50 &59.84	&57.99	&61.04\\
\quad \textit{Emotional Knowledge ($P_e$)} \\
\qquad GPT-4o \;\includegraphics[width=8pt]{openai.png}&
65.52&48.61	&76.39	&76.39	&49.87&68.57	&83.87	&67.03 &64.36 &66.31\\
\qquad Cluade 3.5 Sonnet \;\includegraphics[width=8pt]{Anthropic-AI.png}&
62.27&40.14	&70.73	&80.77	&42.04	&47.39	&86.75 &61.44	&56.62	&61.33\\
\qquad Gemini-2.0-flash \;\includegraphics[height=8pt]{Google_Gemini_logo.svg.png}&
66.54&48.43	&73.47	&77.78	&53.98	&65.93	&79.37 &66.50	&64.92	&66.33\\
\quad \textit{Combined ($P_d + P_c + P_e$)} \\
\qquad GPT-4o \;\includegraphics[width=8pt]{openai.png}&
64.94&51.37	&79.52	&79.77	&44.73	&59.69	&86.75 &66.68	&62.62	&65.74\\
\qquad Cluade 3.5 Sonnet \;\includegraphics[width=8pt]{Anthropic-AI.png}&
67.65&45.45	&76.83	&\textbf{83.64}	&49.53	&58.72	&84.71 &66.65	&63.51	&66.62\\
\qquad Gemini-2.0-flash \;\includegraphics[height=8pt]{Google_Gemini_logo.svg.png}&
65.28&48.18	&72.73	&76.83	&53.71	&66.67	&78.95 &66.05	&64.41	&65.71\\
\midrule
\textbf{\model}\;\includegraphics[height=8pt]{Google_Gemini_logo.svg.png} \;\includegraphics[width=8pt]{Anthropic-AI.png}\;\includegraphics[width=8pt]{openai.png}
&\textbf{72.46}&52.27 	&76.06	&80.50	&\textbf{59.87}	&\textbf{77.37}	&\textbf{90.57}	&\textbf{72.73}	&\textbf{71.15}	&\textbf{72.45}\\
\quad $-$ Aspect-specific Prompting & 70.95	&49.70	&\textbf{80.28}	&82.12	&58.27	&73.91	&87.90 &71.88	&70.31	&71.51\\


\bottomrule
\end{tabular}%
}
\end{table*}

\noindent

\paragraph{Baseline Results}
As shown in Table \ref{tab:benchmarking}, \model\ outperforms the previous State-of-the-Art results of \citep{Yadav-etal-2023-towards} by 7.55\% in macro-F1 score and 7.78\% in weighted-F1 score. The setup by \citet{Yadav-etal-2023-towards} utilizes multimodal fusion based on conditional adaptive gating with pre-trained ResNet and BERT models which are fine-tuned for the task of depressive symptom classification. The superior performance of \model\ highlights the effectiveness of Large Language Model based approaches over these traditional pre-trained models. This is further confirmed compared to the fine-tuned unimodal and multimodal baselines which sit well below the benchmark set by \model. Further, Table \ref{tab:benchmarking} highlights the importance of the textual modality for the fine-grained depressive symptom analysis task as shown by the sharp improvement in performance from the unimodal image setup to the unimodal text setup. The utility of the LLM generated explanations is also established as a consistent increase in performance can be seen by substituting OCR with LLM generated explanations as the textual modality as seen with the BART model improving in macro-F1 from 44.71\% with OCR to 55.81\% with explanations.

\begin{figure}[t]
    \centering
    \includegraphics[width=0.85\columnwidth]{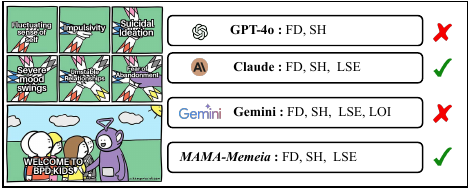}
    \caption{Meme example with predicted labels from GPT-4o, Claude 3.5 Sonnet, Gemini-2.0-flash and \model.}
    \label{fig:qualitative}
\end{figure}

\paragraph{Experimentation with Multimodal LLMs}
We experiment with a variety of different open-source and closed-source models to provide explanations for memes as shown in Table \ref{tab:results}. These setups are compared in both Single-Agent and Multi-Agent setups, as well as with a variety of prompting setups including the Multi-Aspect prompting (cf. Section \ref{sec:multi-aspect}). Finally, we compare \model\ and other LLM based approaches with results derived using gold-label human-written explanations. For a representative variety in the generated content, we put careful consideration into the choice of our models. We take particular effort to utilise a mix of open-source and closed-source models. We also ensure a variety in the base foundational LLMs that these Multimodal LLMs are built upon.
Based on these criterias, we generate explanations from six popular Multimodal LLMs
- \llava\ \citep{liu2023llava}, \llavanext\ \citep{liu2024llavanext}, \mcpm\ \citep{hu2024minicpm}, \gpt\ \citep{openai2024gpt4ocard},  Claude 3.5 Sonnet \cite{claude}, Gemini-2.0-flash \cite{geminiteam2024geminifamilyhighlycapable}.

\paragraph{Results and Ablation}
As seen in Table \ref{tab:results}, the performance of closed-source models such as Claude 3.5 Sonnet is significantly better compared to the open-source models such as LLaVA 1.5. The significant performance boost for the GPT-4o and Gemini-2.0-flash models with the Multi-Aspect Prompting setup indicates towards the effectiveness of these knowledge-based prompting setups. This is also noted with the ablation with removing aspect-specific prompting (referring to the Multi-Aspect setup). Compared to the Human Explanations, the \model\ framework achieves an improvement of more than 8\% and Claude 3.5 Sonnet achieves an improvement of more than 4\% in weighted-F1 score supporting our proposal for utilizing LLM generated explanations as an automated and low-resource alternative to human annotations. 

\paragraph{Qualitative Analysis} We perform a qualitative analysis of the outputs of the GPT-4o, Claude  3.5 Sonnet and Gemini-2.0-flash models along with the \model\ framework to infer trends in predictions made by these models. An example of this analysis is provided in Figure \ref{fig:qualitative}. As seen in Figure \ref{fig:qualitative}, we observe the that the Gemini-2.0-flash model is prone to over-prediction while the GPT-4o model is prone to under-prediction. This behavior is partly explained when observing the length of the explanations generated by these models where the Gemini-2.0-flash model consistently outputs the lengthiest explanations among the set. However, given the black-box nature of these models, further analysis is required for detailed understanding of this phenomenon. This offers an explanation for the improved performance of the \model\ framework where the averaging of the confidence of each of these models yields better results.

Analyzing the patterns in the multiple rounds of debate between the models, we observe that the models frequently correct and influence each other through their detailed explanations. For instance, as in Figure \ref{fig:model}, only one of the models (Claude 3.5 Sonnet) correctly predicts the symptom of Self-Harm and corrects the other two models in their predictions over the rounds of debate. This self-correction tendency stands out as a major strength of \model.

\section{Conclusion}
We introduced \dataset\, a dataset for fine-grained classification of depressive symptoms in memes augmented with ground-truth human explanations and LLM generated explanations for memes. We presented \model, a Multi-Agent Multi-Aspect inferencing framework, grounded in literature in Psychology, for classifying depressive symptoms in multimodal meme data. We demonstrated that \model\ consistently outperformed traditional methods and improved upon single-agent LLM approaches for the task.

While we utilize closed-source models for our \model\ framework, we look forward to developing methods for effective utilization of open-source LLMs in future work given the pressing need for open and transparent research on LLMs in sensitive domains such as mental health.

\section*{Ethics Statement}
This work builds upon the original \texttt{RESTORE} dataset \cite{Yadav-etal-2023-towards}, which contains publicly available meme images collected from social media; the original authors ensured that all data were anonymized. Because our study concerns the sensitive domain of mental health content, we emphasize that the proposed system identifies depressive symptoms expressed within meme content only and is not intended for diagnosing or profiling any individual. Our multi-agent framework leverages large language models that may introduce additional biases, and we release all methods and data resources solely for research purposes. We caution against deploying such LLM-based systems in high-stakes mental health or content-moderation settings without substantial expert human oversight, particularly for critical symptoms such as self-harm. We further acknowledge that the closed-source models used in our framework provide limited transparency regarding their architectures and training data, which constrains interpretability. Finally, because our approach performs inference using existing pretrained models rather than training large models from scratch, its environmental impact is comparatively limited.

\small

\bibliography{aaai2026}


\end{document}